\documentclass[11pt]{article}
\usepackage{acl2016}
\usepackage{times}
\usepackage{latexsym}
\usepackage{url}
\usepackage{algorithm}
\usepackage{algpseudocode}
\usepackage{pifont}
\usepackage{times}
\usepackage{url}
\usepackage{multirow}
\usepackage{amssymb}
\usepackage{amsthm}
\usepackage{amsmath}
\usepackage{verbatim}
\usepackage{tikz-qtree}
\usepackage{tikz}

\algrenewcomment[1]{\(\triangleright\) #1}
\algnewcommand{\LineComment}[1]{\State \(\triangleright\) #1}

\newenvironment{itemizesquish}{\begin{list}{\labelitemi}{\setlength{\itemsep}{-0.2em}\setlength{\labelwidth}{0.5em}\setlength
{\leftmargin}{\labelwidth}\addtolength{\leftmargin}{\labelsep}}}{\end{list}}

\newcommand{\vect}[1]{\mathbf{#1}}

\newcommand{\ignore}[1]{}
\newcommand{\Sref}[1]{\S\ref{#1}}

\newcommand{\tref}[1]{table~\ref{#1}}

\usepackage[T1]{fontenc}

\aclfinalcopy 
\def\aclpaperid{13} 

\title{Correlation-based Intrinsic Evaluation of Word Vector Representations}
\author{Yulia Tsvetkov$^{\spadesuit}$ ~ Manaal Faruqui$^{\spadesuit}$ ~ Chris Dyer$^{\clubsuit\spadesuit}$ \\
 $^\spadesuit$Carnegie Mellon University ~~ $^\clubsuit$Google DeepMind  \\
{ \tt $\{$ytsvetko,mfaruqui,cdyer\}@cs.cmu.edu}}
\date{}

\begin{document}

\maketitle

\begin{abstract}
We introduce \textsc{qvec-cca}---an intrinsic evaluation metric for word vector representations based on correlations of learned vectors  
with features extracted from linguistic resources. 
We show that \textsc{qvec-cca} scores  
are an effective proxy for a range of extrinsic semantic and syntactic tasks. 
We also show that the proposed evaluation obtains higher and more consistent correlations with 
downstream tasks, compared to existing approaches
to intrinsic evaluation of word vectors that are based on word similarity. 

\end{abstract}

\section{Introduction}
Being linguistically opaque, vector-space representations of words---word embeddings---have limited practical value as standalone items.  
They are effective, however, in representing 
meaning---through individual dimensions and combinations 
of thereof---when used as features 
in downstream applications \cite[\textit{inter alia}]{turian:2010,lazaridou:2013,socher:2013,bansal:2014,guo2014revisiting}.  
Thus, unless it is coupled with an extrinsic task, 
intrinsic evaluation of word vectors has little value in itself. 
The main purpose of an intrinsic evaluation is to serve as a 
\textit{proxy} for the downstream task the embeddings 
are tailored for.  
This paper advocates a novel approach to constructing such a proxy. 

What are the desired properties of an intrinsic evaluation measure of word embeddings?  
First, retraining models that use word embeddings as features is often expensive. 
A \textit{computationally efficient} intrinsic evaluation 
that \textit{correlates with extrinsic scores} is useful for faster prototyping. 
Second, an intrinsic evaluation that enables \textit{interpretation} and analysis 
of properties encoded by vector dimensions 
is an auxiliary mechanism for analyzing how these properties affect the 
target downstream task. It thus facilitates refinement of word vector models 
and, consequently, improvement of the target task.  
Finally, an intrinsic evaluation that approximates a range of 
related downstream tasks (e.g., semantic text-classification tasks) 
allows to assess \textit{generality} (or specificity) of a word vector model, 
without actually implementing all the tasks. 

\newcite{qvec:15} proposed an evaluation measure---\textsc{qvec}---that was shown to 
correlate well with downstream semantic tasks. Additionally, it helps shed new light on how vector spaces encode meaning thus facilitating the interpretation of word vectors.
The crux of the method is to correlate distributional word vectors with linguistic word vectors constructed 
from rich linguistic resources, annotated by domain experts. 
\textsc{Qvec} can easily be adjusted to specific downstream tasks (e.g., part-of-speech tagging) by selecting task-specific linguistic resources (e.g., part-of-speech annotations). 
However, \textsc{qvec} suffers from two weaknesses. 
First, it is not invariant to  linear transformations of 
the embeddings' basis, whereas the bases in word embeddings 
are generally arbitrary \cite{szegedy2013intriguing}. 
Second, it produces an unnormalized score: the more dimensions 
in the embedding matrix the higher the score. 
This precludes comparison of models of different dimensionality.
In this paper, we introduce \textsc{qvec-cca}, which simultaneously 
addresses both problems, while preserving major strengths of \textsc{qvec}.\footnote{\url{https://github.com/ytsvetko/qvec}} 

\section{\textsc{qvec} and \textsc{qvec-cca}}
\label{sec:model}
We introduce \textsc{qvec-cca}---an intrinsic evaluation measure of the quality of word embeddings. 
Our method is a modification of \textsc{qvec}---an evaluation based on alignment of embeddings to a matrix of features extracted from a linguistic resource \cite{qvec:15}. 
We review \textsc{qvec}, and then describe \textsc{qvec-cca}.

\paragraph{\textsc{qvec}.} 
The main idea behind \textsc{qvec} is to quantify the linguistic content of word embeddings 
by maximizing the correlation with a manually-annotated linguistic resource. 
Let the number of common words in the vocabulary of the word embeddings and the linguistic resource be $N$. 
To quantify the semantic content of embeddings, a semantic/syntactic linguistic matrix $\vect{S} \in \mathbb{R}^{P \times N}$ is constructed from a semantic/syntactic database, with a column vector for each word. 
Each word vector is a distribution of the word over $P$ linguistic properties, 
based on annotations of the word in the database. 
Let $\vect{X} \in \mathbb{R}^{D \times N}$ be embedding matrix with every row 
as a dimension vector $\vect{x} \in \mathbb{R}^{1 \times N}$. $D$ denotes the 
dimensionality of word embeddings. 
Then, $\vect{S}$ and $\vect{X}$ are aligned to maximize the 
cumulative correlation between the aligned dimensions of the two matrices. 
Specifically, let $\vect{A} \in \{0,1\}^{D \times P}$ be a matrix of alignments 
such that $a_{ij} = 1$ iff $\vect{x}_i$ is aligned to $\vect{s}_j$, otherwise $a_{ij} = 0$. 
If $r(\vect{x}_i, \vect{s}_j)$ is the Pearson's correlation between 
vectors $\vect{x}_i$ and $\vect{s}_j$, then \textsc{qvec} is defined as: 
\begin{align*}
{\normalfont\textsc{qvec}} = \max_{\vect{A} : \sum_{j} a_{ij} \leq 1}
\sum_{i=1}^{X} \sum_{j=1}^{S} r(\vect{x}_i, \vect{s}_j) \times a_{ij}
\end{align*}
The constraint $\sum_{j} a_{ij} \leq 1$, warrants that one distributional 
dimension is aligned to at most one linguistic dimension.

\paragraph{\textsc{qvec-cca}.}
To measure correlation between the embedding matrix $\vect{X}$ and the linguistic matrix $\vect{S}$, 
instead of cumulative dimension-wise correlation we employ canonical correlation analysis \cite[CCA]{hardoon2004canonical}.  
CCA finds two sets of basis vectors, one for $\vect{X^\top}$ and the other for $\vect{S^\top}$, 
such that the correlations between the projections of the matrices onto these basis vectors 
are maximized. 
Formally, CCA finds a pair of basis vectors $\vect{v}$ and $\vect{w}$ such that 
\begin{align*}
\textsc{qvec-cca} &=\mathrm{CCA}(\vect{X^\top}, \vect{S^\top}) \\
                 &= \max_{\vect{v},\vect{w}} r(\vect{X^\top} \vect{v}, \vect{S^\top} \vect{w}) \nonumber
\end{align*} 
Thus, \textsc{qvec-cca} ensures invariance to the matrices' bases' rotation, and since it is a single correlation, it produces a score in $[-1, 1]$.

\section{Linguistic Dimension Word Vectors}
\label{sec:resources}
Both \textsc{qvec} and \textsc{qvec-cca} rely on a matrix of 
linguistic properties constructed from a manually crafted linguistic resource. 
Linguistic resources are invaluable as they capture generalizations made by domain experts.  
However, resource construction is expensive, therefore it is not always 
possible to find an existing resource that captures exactly the set of optimal lexical 
properties for a downstream task. 
Resources that capture more coarse-grained, general properties can be used instead,  
for example, WordNet for semantic evaluation \cite{wordnet}, 
or Penn Treebank \cite[PTB]{marcus1993building} for syntactic evaluation. 
Since these properties are not an exact match to the task, 
the intrinsic evaluation tests for a necessary 
(but possibly not sufficient) set of generalizations.

\paragraph{Semantic vectors.} To evaluate the semantic content of word vectors, 
\newcite{qvec:15} exploit supersense annotations in a WordNet-annotated corpus---SemCor \cite{semcor}. 
The resulting supersense-dimension matrix has 4,199 rows (supersense-annotated nouns and verbs that occur in SemCor at least 5 times\footnote{We exclude sparser word types to avoid skewed probability estimates of senses of polysemous words.}), and 41 columns: 26 for nouns and 15 for verbs. Example vectors are shown in \tref{tab:ling-wn}.   
\begin{table}[!htb]
  \centering
  \small
  \begin{tabular}{lcccc}
  \textbf{\textsc{word}} & \textbf{\textsc{nn.animal}} & \textbf{\textsc{nn.food}} & \textbf{\tiny{$\cdots$}} & \textbf{\textsc{vb.motion}} \\
  \hline
  fish & 0.68 & 0.16 &\tiny{$\cdots$}& 0.00 \\
  duck & 0.31 & 0.00 &\tiny{$\cdots$}& 0.69 \\
  chicken & 0.33 & 0.67 &\tiny{$\cdots$}& 0.00 \\
  \end{tabular}
    \caption{Linguistic dimension word vector matrix with semantic vectors, constructed using SemCor.}
  \label{tab:ling-wn}
\end{table}

\paragraph{Syntactic vectors.} Similar to semantic vectors, we construct syntactic vectors for all words 
with 5 or more occurrences in the training part of the PTB. 
Vector dimensions are probabilities of the part-of-speech (POS) annotations in the corpus. 
This results in 10,865 word vectors with 45 interpretable columns, each column corresponds to a POS tag from the PTB; a snapshot is shown in \tref{tab:ling-ptb}.
\begin{table}[!htb]
  \centering
  \small
  \begin{tabular}{lcccc}
  \textbf{\textsc{word}} & \textbf{\textsc{ptb.nn}} & \textbf{\textsc{ptb.vb}} & \textbf{\tiny{$\cdots$}} & \textbf{\textsc{ptb.jj}} \\
  \hline
  spring & 0.94 & 0.02 &\tiny{$\cdots$}& 0.00 \\
  fall   & 0.49 & 0.43 &\tiny{$\cdots$}& 0.00 \\
  light  & 0.52 & 0.02 &\tiny{$\cdots$}& 0.41 \\
  \end{tabular}
    \caption{Linguistic dimension word vector matrix with syntactic vectors, constructed using PTB.}
  \label{tab:ling-ptb}
\end{table}

\section{Experiments}
\label{sec:experiments}

\paragraph{Experimental setup.} 
We replicate the experimental setup of \newcite{qvec:15}: 
\begin{itemizesquish}
\item We first train 21 word vector models: variants of 
CBOW and Skip-Gram models \cite{mikolov-iclr13};  
their modifications CWindow, Structured Skip-Gram, and CBOW with Attention \cite{wang2vec,wang2vec-attention}; 
GloVe vectors \cite{glove:2014}; 
Latent Semantic Analysis (LSA) based vectors \cite{Church:1990:WAN:89086.89095}; 
and retrofitted GloVe and LSA vectors \cite{faruqui:2015}. 
\item We then evaluate these word vector models using existing \textit{intrinsic} evaluation methods: 
{\sc qvec} and the proposed  {\sc qvec-cca}, and also 
word similarity tasks using the WordSim353 dataset~\cite[WS-353]{citeulike:379845}, 
MEN dataset~\cite{bruni:2012}, 
and SimLex-999 dataset \cite[SimLex]{HillRK14}.\footnote{We employ an implementation of a suite of word similarity tasks at \url{wordvectors.org} \cite{faruqui-2014:SystemDemo}.} 
\item In addition, the same vectors are evaluated using \textit{extrinsic} text classification tasks. Our semantic benchmarks are 
four binary categorization tasks from the 20 Newsgroups (20NG); 
sentiment analysis task \cite[Senti]{socher:2013}; 
and the metaphor detection \cite[Metaphor]{tsvetkov-acl-14}. 
\item Finally, we compute the Pearson's correlation coefficient $r$ to quantify the linear relationship between the intrinsic and extrinsic scorings. The higher the correlation, the better suited the intrinsic evaluation to be used as a proxy to the extrinsic task. 
\end{itemizesquish}

We extend the setup of \newcite{qvec:15} with two syntactic benchmarks, 
and evaluate {\sc qvec-cca} with the syntactic matrix. 
The first task is POS tagging; we use the LSTM-CRF model \cite{lample2016neural}, 
and the second is dependency parsing (Parse), using the stack-LSTM
model of \newcite{dyer2015transition}. 

\paragraph{Results.}
To test the efficiency of {\sc qvec-cca} in capturing the semantic content of word vectors, 
we evaluate how well the scores correspond to the scores of word vector models on semantic benchmarks. 
{\sc Qvec} and {\sc qvec-cca} employ the semantic supersense-dimension vectors described in \Sref{sec:resources}. 
In \tref{tab:sst-eval}, we show correlations between intrinsic scores 
(word similarity/{\sc qvec}/{\sc qvec-cca}) and extrinsic scores across 
semantic benchmarks for 300-dimensional vectors. 
{\sc qvec-cca} obtains high positive correlation with all the semantic tasks, and outperforms {\sc qvec} on two tasks.
\begin{table}[!htb]
  \centering
\begin{tabular}{lccc}
			    &	\textbf{20NG}	& \textbf{Metaphor}		& \textbf{Senti}	 \\ \hline
\bfseries{WS-353}	&	0.55 			& 0.25 				& 0.46 \\
\bfseries{MEN}	&	0.76	& 0.49				& 0.55 \\
\bfseries{SimLex}	&	0.56			& 0.44 				& 0.51 \\\hline
\bfseries{\scshape{qvec}}	& 0.74	& {0.75}		& 0.88 \\ 
\bfseries{\scshape{qvec-cca}}	& {0.77}	& 0.73		& {0.93} \\ 
\hline
\end{tabular}
\caption{Pearson's correlations between word similarity/{\sc qvec}/{\sc qvec-cca} 
scores and the downstream text classification tasks.}
  \label{tab:sst-eval}
\end{table}

In \tref{tab:ptb-eval}, we evaluate {\sc qvec} and {\sc qvec-cca} on syntactic benchmarks. 
We first use linguistic vectors with dimensions corresponding to 
part-of-speech tags (denoted as  \textsc{ptb}).  
Then, we use linguistic vectors which are a concatenation of the 
semantic and syntactic matrices described in \Sref{sec:resources} 
for words that occur in both matrices; this setup is denoted as \textsc{ptb+sst}. 
\begin{table}[!htb]
  \centering
\begin{tabular}{clrr}
		&	    &	\textbf{POS}	& \textbf{Parse} \\ 
\cline{2-4}
&\bfseries{WS-353}											&	-0.38	&  {0.68}	\\
&\bfseries{MEN}												&	-0.32	& 	0.51		 \\
&\bfseries{SimLex}											&	0.20	& 	-0.21		\\
\cline{2-4}
\multirow{2}{*}{\textsc{ptb}}&\bfseries{\scshape{qvec}}	&	0.23	&		0.39	\\ 
&\bfseries{\scshape{qvec-cca}}								&	0.23	&	0.50	\\ 
\cline{2-4}
\multirow{2}{*}{\textsc{ptb+sst}}&\bfseries{\scshape{qvec}}&	{0.28}	&	0.37		\\ 
&\bfseries{\scshape{qvec-cca}}								&0.23		&	0.63		\\ 
\cline{2-4}
\end{tabular}
\caption{Pearson's correlations between word similarity/{\sc qvec}/{\sc qvec-cca} scores and the downstream syntactic tasks.}
  \label{tab:ptb-eval}
\end{table}

Although some word similarity tasks obtain high correlations with syntactic applications, 
these results are inconsistent, and vary from a high negative to a high positive correlation. 
Conversely, {\sc qvec} and {\sc qvec-cca} consistently obtain moderate-to-high positive correlations 
with the downstream tasks. 

Comparing performance of {\sc qvec-cca} in \textsc{ptb} and \textsc{ptb+sst} setups 
sheds light on the importance of linguistic signals captured by the linguistic matrices. 
Appending supersense-annotated columns to the linguistic matrix 
which already contains POS-annotated columns does not affect correlations 
of {\sc qvec-cca} with the POS tagging task, since the additional linguistic information  
is not relevant for approximating how well dimensions of word embeddings encode POS-related properties. 
In the case of dependency parsing---the task which encodes not only syntactic, but also semantic 
information (e.g., captured by subject-verb-object relations)---supersenses 
introduce relevant linguistic signals that are not present in POS-annotated columns. 
Thus, appending supersense-annotated columns to the linguistic matrix improves 
correlation of {\sc qvec-cca} with the dependency parsing task. 
 
\ignore{
\section{Future Work}
\label{sec:proposal}
1) linguistic resource

constructing a linguistic matrix that better reflects features important for downstream tasks, we can build better proxys to downstream tasks.

2) adapting the intrinsic evaluation to non-linear match to a linguistic resource (via transfer learning) 

3) in-depth analysis of interpretation
}

\section{Conclusion}
We introduced {\sc qvec-cca}---an approach to intrinsic evaluation of 
word embeddings. We also showed that both {\sc qvec} and {\sc qvec-cca} 
are not limited to semantic evaluation, but are general approaches, 
that can evaluate word vector content with respect to desired linguistic properties. Semantic and syntactic linguistic features that we use to 
construct linguistic dimension matrices are rather coarse, thus the proposed evaluation can approximate a range of downstream tasks, but may not be sufficient to evaluate finer-grained features. 
In the future work we propose to exploit existing semantic, syntactic, morphological, and typological resources (e.g., universal
dependencies treebank \cite{agic:2015} and WALS \cite{wals2013}), and also multilingual resources (e.g., Danish supersenses \cite{martinezalonsoetal2015supersenses}) to construct better linguistic matrices, suited for evaluating vectors used in additional NLP tasks.  

\section*{Acknowledgments}
This work was supported by the National Science Foundation through award IIS-1526745. We thank Benjamin Wilson for helpful comments. 


\bibliography{qvec_cca}

\begin{thebibliography}{}

\bibitem[\protect\citename{Agi{\'c} \bgroup et al.\egroup }2015]{agic:2015}
{\v Z}eljko Agi{\'c}, Maria~Jesus Aranzabe, Aitziber Atutxa, Cristina Bosco,
  Jinho Choi, Marie-Catherine de~Marneffe, Timothy Dozat, Rich{\'a}rd Farkas,
  Jennifer Foster, Filip Ginter, Iakes Goenaga, Koldo Gojenola, Yoav Goldberg,
  Jan Haji{\v c}, Anders~Tr{\ae}rup Johannsen, Jenna Kanerva, Juha Kuokkala,
  Veronika Laippala, Alessandro Lenci, Krister Lind{\'e}n, Nikola Ljube{\v
  s}i{\'c}, Teresa Lynn, Christopher Manning, H{\'e}ctor~Alonso Mart{\'i}nez,
  Ryan {McDonald}, Anna Missil{\"a}, Simonetta Montemagni, Joakim Nivre, Hanna
  Nurmi, Petya Osenova, Slav Petrov, Jussi Piitulainen, Barbara Plank, Prokopis
  Prokopidis, Sampo Pyysalo, Wolfgang Seeker, Mojgan Seraji, Natalia Silveira,
  Maria Simi, Kiril Simov, Aaron Smith, Reut Tsarfaty, Veronika Vincze, and
  Daniel Zeman.
\newblock 2015.
\newblock Universal dependencies 1.1.
\newblock {LINDAT}/{CLARIN} digital library at Institute of Formal and Applied
  Linguistics, Charles University in Prague.

\bibitem[\protect\citename{Bansal \bgroup et al.\egroup }2014]{bansal:2014}
Mohit Bansal, Kevin Gimpel, and Karen Livescu.
\newblock 2014.
\newblock Tailoring continuous word representations for dependency parsing.
\newblock In {\em Proc. of ACL}.

\bibitem[\protect\citename{Bruni \bgroup et al.\egroup }2012]{bruni:2012}
Elia Bruni, Gemma Boleda, Marco Baroni, and Nam-Khanh Tran.
\newblock 2012.
\newblock Distributional semantics in technicolor.
\newblock In {\em Proc. of ACL}.

\bibitem[\protect\citename{Church and Hanks}1990]{Church:1990:WAN:89086.89095}
Kenneth~Ward Church and Patrick Hanks.
\newblock 1990.
\newblock Word association norms, mutual information, and lexicography.
\newblock {\em Computational Linguistics}, 16(1):22--29.

\bibitem[\protect\citename{Dryer and Haspelmath}2013]{wals2013}
Matthew~S. Dryer and Martin Haspelmath, editors.
\newblock 2013.
\newblock {\em WALS Online}.
\newblock Max Planck Institute for Evolutionary Anthropology.
\newblock \url{http://wals.info/}.

\bibitem[\protect\citename{Dyer \bgroup et al.\egroup
  }2015]{dyer2015transition}
Chris Dyer, Miguel Ballesteros, Wang Ling, Austin Matthews, and Noah~A. Smith.
\newblock 2015.
\newblock Transition-based dependency parsing with stack long short-term
  memory.
\newblock In {\em Proc. of ACL}.

\bibitem[\protect\citename{Faruqui and Dyer}2014]{faruqui-2014:SystemDemo}
Manaal Faruqui and Chris Dyer.
\newblock 2014.
\newblock Community evaluation and exchange of word vectors at wordvectors.org.
\newblock In {\em Proc. of ACL (Demonstrations)}.

\bibitem[\protect\citename{Faruqui \bgroup et al.\egroup }2015]{faruqui:2015}
Manaal Faruqui, Jesse Dodge, Sujay~Kumar Jauhar, Chris Dyer, Noah~A. Smith, and
  Eduard Hovy.
\newblock 2015.
\newblock Retrofitting word vectors to semantic lexicons.
\newblock In {\em Proc. of NAACL}.

\bibitem[\protect\citename{Fellbaum}1998]{wordnet}
Christiane Fellbaum, editor.
\newblock 1998.
\newblock {\em {WordNet}: an electronic lexical database}.
\newblock {MIT} Press.

\bibitem[\protect\citename{Finkelstein \bgroup et al.\egroup
  }2001]{citeulike:379845}
Lev Finkelstein, Evgeniy Gabrilovich, Yossi Matias, Ehud Rivlin, Zach Solan,
  Gadi Wolfman, and Eytan Ruppin.
\newblock 2001.
\newblock {Placing search in context: the concept revisited}.
\newblock In {\em Proc. of WWW}.

\bibitem[\protect\citename{Guo \bgroup et al.\egroup }2014]{guo2014revisiting}
Jiang Guo, Wanxiang Che, Haifeng Wang, and Ting Liu.
\newblock 2014.
\newblock Revisiting embedding features for simple semi-supervised learning.
\newblock In {\em Proc. of EMNLP}.

\bibitem[\protect\citename{Hardoon \bgroup et al.\egroup
  }2004]{hardoon2004canonical}
David~R Hardoon, Sandor Szedmak, and John Shawe-Taylor.
\newblock 2004.
\newblock Canonical correlation analysis: An overview with application to
  learning methods.
\newblock {\em Neural Computation}, 16(12):2639--2664.

\bibitem[\protect\citename{Hill \bgroup et al.\egroup }2014]{HillRK14}
Felix Hill, Roi Reichart, and Anna Korhonen.
\newblock 2014.
\newblock {SimLex-999}: Evaluating semantic models with (genuine) similarity
  estimation.
\newblock {\em CoRR}, abs/1408.3456.

\bibitem[\protect\citename{Lample \bgroup et al.\egroup
  }2016]{lample2016neural}
Guillaume Lample, Miguel Ballesteros, Sandeep Subramanian, Kazuya Kawakami, and
  Chris Dyer.
\newblock 2016.
\newblock Neural architectures for named entity recognition.
\newblock In {\em Proc. of NAACL}.

\bibitem[\protect\citename{Lazaridou \bgroup et al.\egroup
  }2013]{lazaridou:2013}
Angeliki Lazaridou, Eva~Maria Vecchi, and Marco Baroni.
\newblock 2013.
\newblock Fish transporters and miracle homes: How compositional distributional
  semantics can help {NP} parsing.
\newblock In {\em Proc. of EMNLP}.

\bibitem[\protect\citename{Ling \bgroup et al.\egroup
  }2015a]{wang2vec-attention}
Wang Ling, Lin Chu-Cheng, Yulia Tsvetkov, Silvio Amir, Ramon Fermandez, Chris
  Dyer, Alan~W Black, and Isabel Trancoso.
\newblock 2015a.
\newblock Not all contexts are created equal: Better word representations with
  variable attention.
\newblock In {\em Proc. of EMNLP}.

\bibitem[\protect\citename{Ling \bgroup et al.\egroup }2015b]{wang2vec}
Wang Ling, Chris Dyer, Alan Black, and Isabel Trancoso.
\newblock 2015b.
\newblock Two/too simple adaptations of \texttt{word2vec} for syntax problems.
\newblock In {\em Proc. of NAACL}.

\bibitem[\protect\citename{Marcus \bgroup et al.\egroup
  }1993]{marcus1993building}
Mitchell~P Marcus, Mary~Ann Marcinkiewicz, and Beatrice Santorini.
\newblock 1993.
\newblock Building a large annotated corpus of {E}nglish: The penn treebank.
\newblock {\em Computational Linguistics}, 19(2):313--330.

\bibitem[\protect\citename{Mart{\'i}nez~Alonso \bgroup et al.\egroup
  }2015]{martinezalonsoetal2015supersenses}
H{\'e}ctor Mart{\'i}nez~Alonso, Anders Johannsen, Sussi Olsen, Sanni Nimb,
  Nicolai~Hartvig S{\o}rensen, Anna Braasch, Anders S{\o}gaard, and
  Bolette~Sandford Pedersen.
\newblock 2015.
\newblock Supersense tagging for {Danish}.
\newblock In {\em Proc. of NODALIDA}, page~21.

\bibitem[\protect\citename{Mikolov \bgroup et al.\egroup }2013]{mikolov-iclr13}
Tomas Mikolov, Kai Chen, Greg Corrado, and Jeffrey Dean.
\newblock 2013.
\newblock Efficient estimation of word representations in vector space.
\newblock In {\em Proc. of ICLR}.

\bibitem[\protect\citename{Miller \bgroup et al.\egroup }1993]{semcor}
George~A. Miller, Claudia Leacock, Randee Tengi, and Ross~T. Bunker.
\newblock 1993.
\newblock A semantic concordance.
\newblock In {\em Proc. of {HLT}}, pages 303--308.

\bibitem[\protect\citename{Pennington \bgroup et al.\egroup }2014]{glove:2014}
Jeffrey Pennington, Richard Socher, and Christopher~D. Manning.
\newblock 2014.
\newblock {GloVe}: Global vectors for word representation.
\newblock In {\em Proc. of EMNLP}.

\bibitem[\protect\citename{Socher \bgroup et al.\egroup }2013]{socher:2013}
Richard Socher, Alex Perelygin, Jean Wu, Jason Chuang, Christopher~D. Manning,
  Andrew~Y. Ng, and Christopher Potts.
\newblock 2013.
\newblock Recursive deep models for semantic compositionality over a sentiment
  treebank.
\newblock In {\em Proc. of EMNLP}.

\bibitem[\protect\citename{Szegedy \bgroup et al.\egroup
  }2014]{szegedy2013intriguing}
Christian Szegedy, Wojciech Zaremba, Ilya Sutskever, Joan Bruna, Dumitru Erhan,
  Ian Goodfellow, and Rob Fergus.
\newblock 2014.
\newblock Intriguing properties of neural networks.
\newblock In {\em Proc. of ICLR}.

\bibitem[\protect\citename{Tsvetkov \bgroup et al.\egroup
  }2014]{tsvetkov-acl-14}
Yulia Tsvetkov, Leonid Boytsov, Anatole Gershman, Eric Nyberg, and Chris Dyer.
\newblock 2014.
\newblock Metaphor detection with cross-lingual model transfer.
\newblock In {\em Proc. of ACL}, pages 248--258.

\bibitem[\protect\citename{Tsvetkov \bgroup et al.\egroup }2015]{qvec:15}
Yulia Tsvetkov, Manaal Faruqui, Wang Ling, Guillaume Lample, and Chris Dyer.
\newblock 2015.
\newblock Evaluation of word vector representations by subspace alignment.
\newblock In {\em Proc. of EMNLP}, pages 2049--2054.

\bibitem[\protect\citename{Turian \bgroup et al.\egroup }2010]{turian:2010}
Joseph Turian, Lev Ratinov, and Yoshua Bengio.
\newblock 2010.
\newblock Word representations: a simple and general method for semi-supervised
  learning.
\newblock In {\em Proc. of ACL}.

\end{thebibliography}
\bibliographystyle{acl2016}
\end{document}